# Predictive Maintenance: Deep Learning-Based Remaining Useful Life Prediction for Combat Aircraft Engines

## Kestirimci Bakım: Savaş Uçağı Motorları için Derin Öğrenme Tabanlı Kalan Kullanım Ömrü Tahmini

Fatih Ürgen[1*], Doğay Altınel[2]

*[1] Istanbul Technical University, Faculty of Science, Department of Physics Engineering, 34467 Istanbul, Türkiye,*
*[2] Istanbul Medeniyet University, Faculty of Engineering and Natural Sciences, Department of Electrical and Electronics Engineering, 34700 Istanbul, Türkiye*

## Abstract

To improve the operational readiness of combat aircraft engines and reduce unplanned maintenance costs, accurately estimating the remaining useful life (RUL) is critical. Traditional maintenance often proves insufficient under dynamic mission profiles. In this study, a deep learning-based predictive maintenance model capable of autonomously extracting features from multivariate sensor data was developed. Using the NASA C-MAPSS FD001 and FD004 datasets, data were converted into sequential blocks via 50- and 30-step sliding windows, respectively. The model's architectural superiority in autonomously extracting temporal degradation features was validated against RF, CNN-LSTM, and BiLSTM baselines. On FD001, it achieved an R-squared ($R^2$) of 0.8901, a 13.28 RMSE, and a 320.34 NASA risk score, demonstrating generalizability on the multi-regime FD004 dataset with a 15.71 RMSE. The proposed maintenance protocol achieved a 0.9973 AUC at the critical 30-cycle threshold, ensuring high reliability. Additionally, a decision-support simulator has been developed to validate this protocol under aggressive combat flight profiles.

## Öz

Savaş uçağı motorlarının operasyonel hazırlık seviyesini artırmak ve planlanmamış bakım maliyetlerini düşürmek için, kalan faydalı ömrün (RUL) doğru tahmini kritiktir. Geleneksel bakımlar, dinamik görev profillerinde çalışan uçaklar için genellikle yetersiz kalmaktadır. Bu çalışmada, çok değişkenli sensör verilerinden otonom özellik çıkarabilen derin öğrenme tabanlı bir kestirimci bakım modeli geliştirilmiştir. NASA C-MAPSS FD001 ve FD004 veri setleri kullanılarak, veriler sırasıyla 50 ve 30 adımlık bir kayan pencere yöntemiyle ardışık bloklara dönüştürülmüştür. Zamansal aşınma özelliklerini otonom olarak çıkarmak amacıyla tasarlanan mimari üstünlüğü; RF, CNN-LSTM ve BiLSTM modelleriyle kıyaslanarak doğrulanmıştır. Geliştirilen model; FD001 üzerinde 0.8901 R-kare ($R^2$), 13.28 RMSE ve 320.34 NASA asimetrik risk skoru elde etmiş; çoklu-rejimli FD004 setinde ise 15.71 RMSE ile genellenebilirliğini kanıtlamıştır. Önerilen modele dayalı bakım protokolü, kritik 30 uçuş döngüsü eşiğinde 0.9973 AUC değerine ulaşarak yüksek güvenilirlik sağlamıştır. Ayrıca, protokolü agresif muharebe uçuş profilleri altında doğrulamak için bir karar-destek simülatörü geliştirilmiştir.

*Corresponding Author
E-mail: urgen25@itu.edu.tr

## 1. INTRODUCTION

The aviation industry is a sector with extremely low fault tolerance, where passenger safety and operational continuity must be maintained at the highest standards. Turbofan engines are among the most critical and costly components within this ecosystem. Currently, airlines and maintenance, repair, and overhaul (MRO) organizations typically employ a time-based maintenance strategy based on fixed flight hours or cycles determined by manufacturers when planning engine maintenance. However, this traditional approach ignores the actual environmental conditions and the dynamic wear processes that the engine is exposed to during operations. Fixed-period maintenance causes expensive engine parts that have not yet completed their useful life to be discarded prematurely, leading to significant economic waste. Furthermore, it fails to detect components that degrade much faster than expected due to aggressive mission profiles, thereby posing a catastrophic safety risk. To overcome these critical problems, the concept of predictive maintenance (PdM), driven by Industry 4.0 paradigms and advanced data analytics, has come to the fore in recent years [1,2].

Predictive maintenance assesses the instantaneous health state of the machine by analyzing real-time data flowing from hundreds of sensors (e.g., temperature, pressure, fan speed) on the engine and predicts in advance when it will fail. In the literature, this task is commonly referred to as remaining useful life (RUL) estimation. Recently, advances in artificial intelligence, particularly in deep learning, have enabled the design of autonomous systems capable of learning degradation patterns in complex and noisy time-series sensor data without requiring manual feature engineering [3]. In this context, Saxena et al. established the foundation for RUL estimation research by introducing the NASA commercial modular aero-propulsion system simulation (C-MAPSS) dataset, which simulates aircraft engine run-to-failure degradation [4].

Building upon the foundational C-MAPSS dataset, early deep learning approaches like Babu et al. demonstrated the advantages of convolutional neural network (CNN) for automatic feature extraction [5]. However, because engine degradation is inherently a time-series problem, Zheng et al. successfully applied long short-term memory (LSTM) networks to capture cumulative wear trends, proving its superiority over traditional methods [6]. Recent studies have heavily focused on increasing algorithmic complexity to improve pure prediction accuracy. For instance, the literature has seen a surge in highly intricate hybrid architectures [7], [8] and multi-scale attention-based bidirectional long short-term memory (BiLSTM) frameworks [9]. Moreover, to efficiently capture long-range temporal dependencies and complex cross-sensor correlations, recent studies have increasingly adopted transformer-based architectures and advanced self-attention mechanisms [10-12]. By dynamically weighting the significance of critical degradation features across an engine's entire operational history, these models overcome the sequential processing bottlenecks of traditional recurrent networks, establishing new benchmarks in pure predictive accuracy. While these existing studies are beneficial for mathematical optimization on static datasets, their substantial computational overhead and architectural complexity often hinder their integration into real-world operational scenarios particularly the real-time what-if simulations required for dynamic and aggressive flight profiles of military combat aircraft. Unlike these purely algorithmic studies, our proposed model bridges the critical gap between data science and operational aviation. Instead of merely calculating RUL, this study integrates a robust, computationally efficient, and streamlined LSTM architecture into a novel decision-support what-if flight simulator capable of executing real-time inferences under simulated combat stress.

Furthermore, to achieve a comprehensive and methodologically sound validation environment, this study evaluates the proposed architecture across two distinct operational boundaries by utilizing both the FD001 and FD004 sub-datasets. Because the baseline FD001 dataset represents a single fault mode under unvarying operational conditions, it provides an uncorrupted and stable environment. Establishing a highly accurate baseline on FD001 is a fundamental prerequisite for this research, as it allows the what-if simulation framework to mathematically inject controlled combat stress multipliers and isolate the model's response to simulated aggressive maneuvers without confounding environmental variables. Subsequently, extending the evaluation to the multi-condition FD004 dataset exposes the exact same prognostic architecture to severe operational noise, six distinct flight regimes, and two simultaneous fault modes. This dual-dataset strategy ensures that the framework's baseline predictive behavior is verified in isolation on FD001, while its cross-environment generalizability, robustness, and capacity to handle severe operational variations are rigorously validated on FD004. Additionally, highlighting the growing trust in these architectures, recent studies have confirmed that data-driven autonomous models, specifically LSTMs, are highly reliable for modeling flight phases [13] and estimating critical aerospace data [14]. To systematically address these operational challenges, this study presents a comprehensive prognostic framework tailored for military aviation. The proposed workflow transitions from conventional predictive maintenance strategies towards advanced deep learning methodologies. While recent literature heavily explores complex CNN or BiLSTM variants, this research adopts a streamlined, pure LSTM architecture utilizing a multivariate sliding-window approach on raw sensor data to successfully capture complex degradation signatures without the need for manual signal smoothing [15]. This choice ensures the optimal balance between capturing the sequential memory of cumulative sensor degradation and maintaining the computational efficiency required for real-time simulator integration. The prognostic performance of the framework is comprehensively evaluated not only through standard statistical metrics like root mean squared error (RMSE) and R-squared ($R^2$), but crucially, through the domain-specific NASA asymmetric risk score [16], ensuring the system adheres to the fail-safe operational doctrines of the aviation industry.

In order to validate the necessity of deep learning for prognostics [17], the proposed architecture is initially benchmarked against a traditional ensemble machine learning baseline, random forest (RF) [18]. While classical algorithms like RF have been widely utilized in predictive maintenance due to their robustness against overfitting, they inherently struggle to natively capture the complex, long-term temporal dependencies present in sequential turbofan degradation trajectories. Additionally, to provide a comprehensive comparative analysis, the proposed model is evaluated against convolutional neural network–long short-term memory (CNN-LSTM) and BiLSTM networks. By comparing our temporal LSTM architecture against these baselines, this study quantitatively evaluates the performance and safety-critical gains achieved over traditional machine learning, as well as its operational stability and computational efficiency compared to other deep learning variants. Based on the proposed model, the maintenance protocol demonstrates reliability in failure detection, effectively maintaining fleet readiness.

## 2. METHODOLOGY

The end-to-end pipeline encompasses the retrieval of raw sensor data, exploratory data analysis (EDA), data transformation, and the architectural design of the deep learning model, concluding with the performance evaluation. All computational processes, including model training and evaluation, were executed on the Google Colab platform utilizing cloud-based GPUs (NVIDIA Tesla T4) to accelerate tensor operations. The entire infrastructure was built using the Python programming language, heavily relying on the TensorFlow and Keras libraries for deep learning frameworks, as well as Pandas and NumPy for data manipulation. The end-to-end workflow begins with the sourcing of the NASA C-MAPSS repository, specifically utilizing the FD001 and FD004 sub-datasets to evaluate the model under varying operational complexities. This is followed by a data preprocessing pipeline, which includes feature normalization, the elimination of zero-variance variables for single-regime data, and the application of a sliding window technique to format the sequential inputs. For multi-regime data, a clustering-based standardization approach is integrated. The processed 3D tensors are then fed into the deep learning architecture, where sequential LSTM layers capture the complex temporal degradation dependencies through backpropagation. Finally, the model outputs the continuous RUL estimation, which is subsequently evaluated against benchmark models using predefined performance metrics.

### 2.1. Dataset Description

The empirical data utilized in this study is sourced from the C-MAPSS dataset, developed and provided by the NASA Ames Prognostics Center of Excellence. To test the validation and generalizability of the proposed framework, two distinct subsets with different levels of complexity were selected, namely FD001 and FD004.

The FD001 subset represents a controlled benchmark scenario operating under a single operational condition at sea level and incorporating a single fault mode characterized by high-pressure compressor degradation. It consists of 100 independent training engines and 100 test engines. In contrast, the FD004 subset represents a highly complex and noisy operational environment that incorporates six distinct flight regimes (varying in altitude, mach number, and throttle resolver angles) and two simultaneous fault modes, which include both high-pressure compressor and fan degradation. The FD004 dataset comprises 249 training engines and 248 test engines. Each recorded instance across both datasets represents a single flight cycle, capturing the engine identification number, the current time cycle, 3 distinct operational settings, and continuous measurements from 21 physical sensors.

## 2.2. Data Preprocessing and Cleaning

Raw sensor data often contains redundant or uninformative variables that can negatively impact the learning efficiency of neural networks. Upon detailed examination of the baseline FD001 dataset, it was identified that settings 1–3 and sensors 1, 5, 10, 16, 18, and 19 exhibited zero variance. To prevent the processing of meaningless information and reduce computational burden, these non-functional variables were completely removed, optimizing the feature space to 15 active variables (Table 1). In contrast, the multi-regime FD004 dataset required the retention of the complete 24-dimensional feature space to preserve environmental context across varying flight conditions.

**Table 1.** Descriptions and selection status of operational settings and sensors based on variance analysis for the baseline FD001 dataset.

| Index | Symbol | Description | Status in Model |
|---|---|---|---|
| Setting 1 | alt | Altitude | Dropped |
| Setting 2 | Mach | Mach Number | Dropped |
| Setting 3 | TRA | Throttle Resolver Angle | Dropped |
| Sensor 1 | T2 | Total temperature at fan inlet | Dropped |
| Sensor 2 | T24 | Total temperature at LPC outlet | Retained |
| Sensor 3 | T30 | Total temperature at HPC outlet | Retained |
| Sensor 4 | T50 | Total temperature at LPT outlet | Retained |
| Sensor 5 | P2 | Pressure at fan inlet | Dropped |
| Sensor 6 | P15 | Total pressure in bypass-duct | Retained |
| Sensor 7 | P30 | Total pressure at HPC outlet | Retained |
| Sensor 8 | Nf | Physical fan speed | Retained |
| Sensor 9 | Nc | Physical core speed | Retained |
| Sensor 10 | epr | Engine pressure ratio (P50/P2) | Dropped |
| Sensor 11 | Ps30 | Static pressure at HPC outlet | Retained |
| Sensor 12 | phi | Ratio of fuel flow to Ps30 | Retained |
| Sensor 13 | NRf | Corrected fan speed | Retained |
| Sensor 14 | NRc | Corrected core speed | Retained |
| Sensor 15 | BPR | Bypass Ratio | Retained |
| Sensor 16 | farB | Burner fuel-air ratio | Dropped |

| Sensor 17 | htBleed | Bleed Enthalpy | Retained |
|---|---|---|---|
| Sensor 18 | Nf_dmd | Demanded fan speed | Dropped |
| Sensor 19 | PCNfR_dmd | Demanded corrected fan speed | Dropped |
| Sensor 20 | W31 | HPT coolant bleed | Retained |
| Sensor 21 | W32 | LPT coolant bleed | Retained |

To validate the feature selection process for FD001, a Pearson correlation analysis was conducted against the target RUL (Figure 1). The retained sensors exhibited strong negative (dark blue) or positive (dark red) correlations, confirming that they carry robust prognostic signals, thereby effectively justifying their inclusion as inputs for the LSTM network. Conversely, since the entirety of the 24-dimensional feature space was retained for the multi-regime FD004 dataset to capture complex operational dynamics, a supplementary correlation heatmap analysis was not required for feature selection in that specific context. To identify the most significant degradation indicators, a Pearson correlation analysis was conducted between the sensor measurements and the RUL (Figure 1). As indicated by the analysis, sensors 11, 4, and 15 exhibited the strongest negative correlations, with exact coefficients of -0.78, -0.76, and -0.72, respectively. This indicates a consistent upward trend in these sensor readings as the engine degrades. In contrast, sensors 12, 7, and 21 demonstrated the strongest positive correlations, with coefficients of 0.75, 0.73, and 0.71, respectively, reflecting a downward trend over the engine's operational life. These precise statistical findings strongly support the inclusion of these specific features in the sequential modeling process. It should be noted that previously eliminated zero-variance features were excluded from this visualization, as the Pearson correlation coefficient is mathematically undefined for constant variables, rendering their inclusion uninformative.

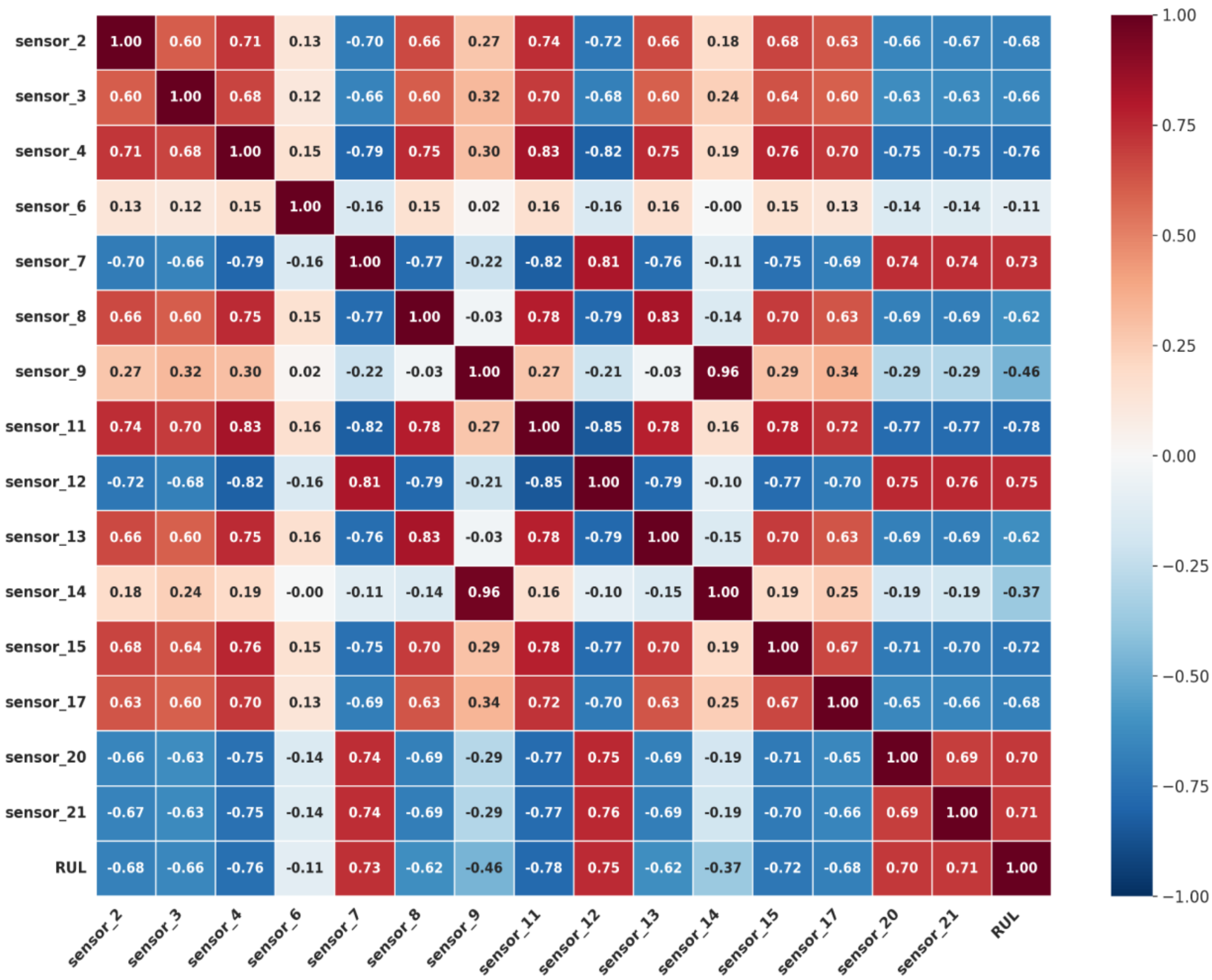

**Figure 1.** Sensor-to-RUL Pearson correlation heatmap for the NASA C-MAPSS FD001 training dataset, illustrating the most significant degradation indicators.

To visually examine the underlying degradation process of the turbofan engines, it is essential to monitor how physical sensor readings evolve over consecutive flight cycles. Figure 2 illustrates the lifecycle behavior of a highly correlated feature, sensor 11, plotted alongside the calculated RUL for a sample unit. During the initial flight cycles, the sensor readings exhibit fluctuations around a stable baseline, representing the healthy operational state of the engine. Correspondingly, the RUL is capped at 125 cycles following the piecewise linear degradation assumption. As the engine progressively deteriorates and approaches failure, a distinct upward trend in the sensor data emerges, inversely mirroring the linear decline in the RUL. This clear temporal dependency demonstrates the necessity of utilizing sequential deep learning architectures, such as LSTM, to capture these complex degradation signatures effectively.

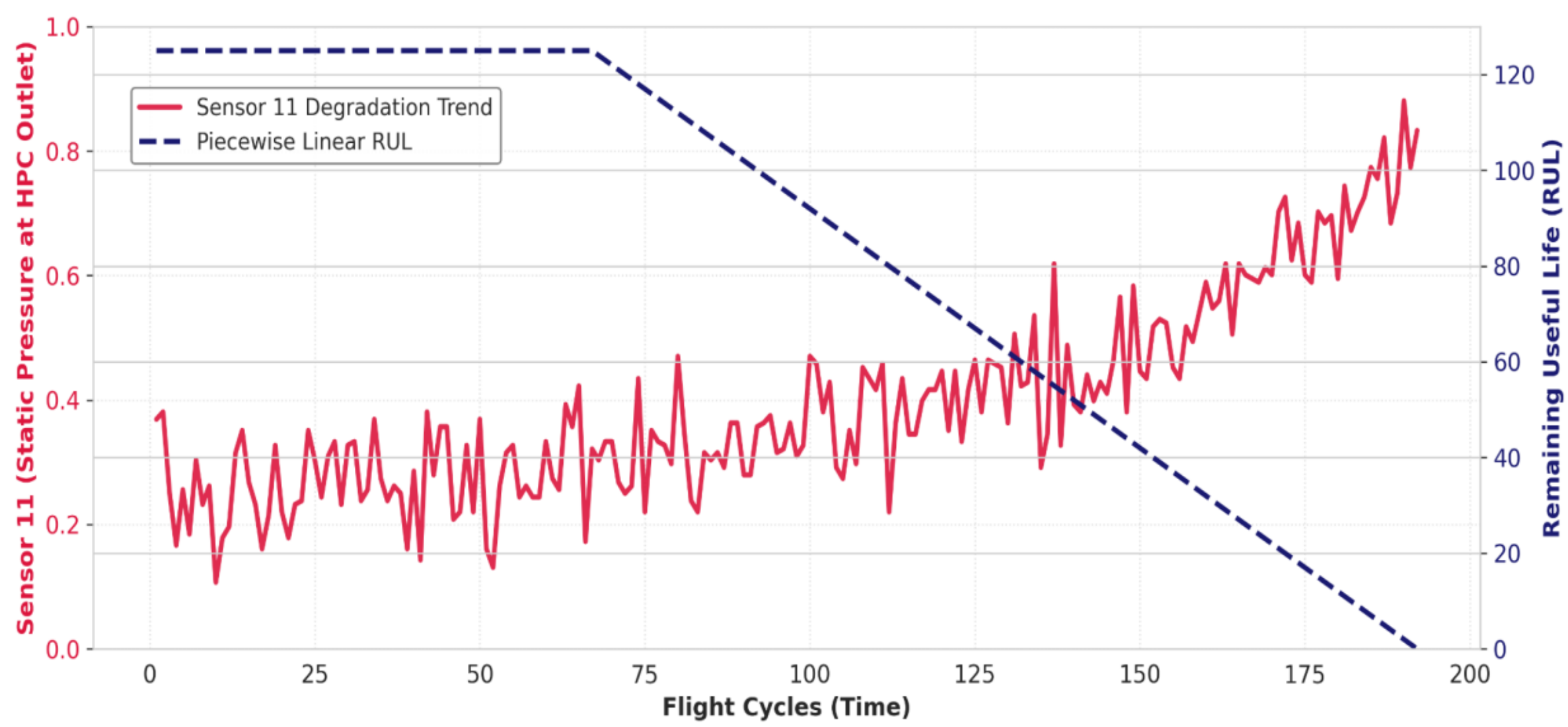

**Figure 2.** Degradation trend of a highly correlated feature (sensor 11) against the piecewise linear RUL trajectory for a sample test engine.

Following feature selection, it is crucial to normalize the data to prevent gradient exploding and to ensure faster convergence during the backpropagation process of the deep learning model. As highlighted in recent aviation machine learning studies [19], proper scaling of complex flight and sensor data is essential for LSTM architectures to function optimally. To ensure that all sensor features contribute equally to the model training and to accelerate convergence, the raw data is scaled to the [0, 1] range using the min-max normalization method [20], as defined in Eq. (1):

$$x_{norm} = \frac{x - x_{min}}{x_{max} - x_{min}} \quad (1)$$

where $x_{norm}$ represents the resulting normalized value, x represents the original raw value of a specific sensor or operational setting, while $x_{min}$ and $x_{max}$ denote the minimum and maximum values of that particular feature observed strictly across the training dataset. To minimize any potential data leakage, the min-max normalization parameters were computed exclusively using the training set. Subsequently, these exact scaling parameters were applied to transform the unseen test dataset.

While this global min-max scaling approach is sufficient for single-regime operations such as the FD001 dataset, datasets with multiple operating conditions require an adapted strategy. In the FD004 dataset, which incorporates six distinct flight regimes and two simultaneous fault modes, global normalization can obscure underlying degradation trends by treating operational shifts (such as changes in altitude, Mach number, or throttle resolver angle) as structural anomalies. Furthermore, unlike the static conditions of FD001 highlighted in Table 1, the operational settings and all 21 sensors exhibit continuous variations across the different flight conditions in FD004. Consequently, no features were dropped during the initial variance analysis for the multi-regime dataset, and the complete 24-dimensional feature space was retained to preserve the full environmental context.

To isolate the environmental noise from the actual mechanical wear, a regime-aware preprocessing pipeline was introduced for the FD004 analysis. As highlighted in recent literature addressing aleatoric uncertainty and operational noise in C-MAPSS datasets [21], handling high variance across multiple flight conditions is critical for robust deep learning predictions. Accordingly, the K-Means clustering algorithm was first deployed to partition the operational space into six discrete clusters representing the distinct flight regimes based on Settings 1–3. Following the clustering phase, the feature variables were standardized using a localized Z-score normalization method applied strictly within each operational regime cluster, as defined in Eq. (2):

$$x_{std} = \frac{x-\mu_r}{\sigma_r} \tag{2}$$

where $x_{std}$ represents the resulting standardized feature value, x denotes the original raw sensor or operational setting value, and $\mu_r$ and $\sigma_r$ represent the mean and standard deviation of that specific feature calculated exclusively within the corresponding operational regime cluster of the training dataset. Both the K-Means cluster boundaries and the localized parameters were computed solely from the training partition and subsequently utilized to transform the test partition.

Furthermore, to ensure full reproducibility and transparency, the data splitting process followed the predefined training and test trajectories of the standard NASA C-MAPSS dataset. The sliding window formatting was applied to these sets in absolute isolation, utilizing a sequence length of 50 for FD001 and an optimized sequence length of 30 for the noisier FD004 environment. Because the designed deep learning model possesses a robust internal capability for autonomous feature extraction, external signal smoothing techniques, such as moving average filters, were intentionally avoided to preserve the raw sequence of sensor degradation data.

## 2.3. Target Labeling

To train the model using a supervised learning paradigm, a definitive target variable representing the RUL had to be established for every single time step of the engine's operation. This study adopted the widely accepted piecewise linear degradation model [22]. The fundamental assumption of this approach is that an engine operates in a perfectly healthy state during its initial cycles, meaning initial wear does not immediately reduce its overall lifespan. Consequently, the maximum RUL is capped at an upper threshold of 125 cycles. The RUL remains constant at 125 during the early life of the engine and only begins to decrease linearly as the engine accumulates wear and approaches its failure point.

## 2.4. Time Series Formatting

Recurrent neural networks, particularly LSTMs, require sequential data to effectively learn cumulative historical dependencies. To accommodate this, the 2-dimensional tabular dataset was transformed into a 3-dimensional tensor structure using the sliding window technique, as schematically represented for the baseline FD001 dataset in Figure 3. For FD001, the length of the window (sequence_length) was empirically determined and set to 50 time steps. This transformation converted the multivariate temporal data of each engine into sequential blocks, allowing the model to look back at the previous 50 flights to make a single prediction. As a result of this operation, the FD001 training data was reshaped into an input matrix with the dimensions [15731, 50, 15], corresponding to the number of samples, time steps, and feature count, respectively. Conversely, for the highly volatile multi-regime FD004 dataset, where engines may exhibit critical failures at earlier flight cycles due to severe operational stress, utilizing a 50-step window would result in padding-induced data distortion for early-failing units. Therefore, the sequence length for FD004 was optimized to 30 time steps. The exact same sliding window methodology was applied to both datasets in absolute isolation to preserve the raw sequence of sensor degradation data without the need for external signal smoothing techniques. Specifically, applying this transformation to the FD004 training dataset reshaped it into an input matrix with dimensions of [61249, 30, 24], reflecting the optimized 30-step window and the retention of the full 24-dimensional feature space.

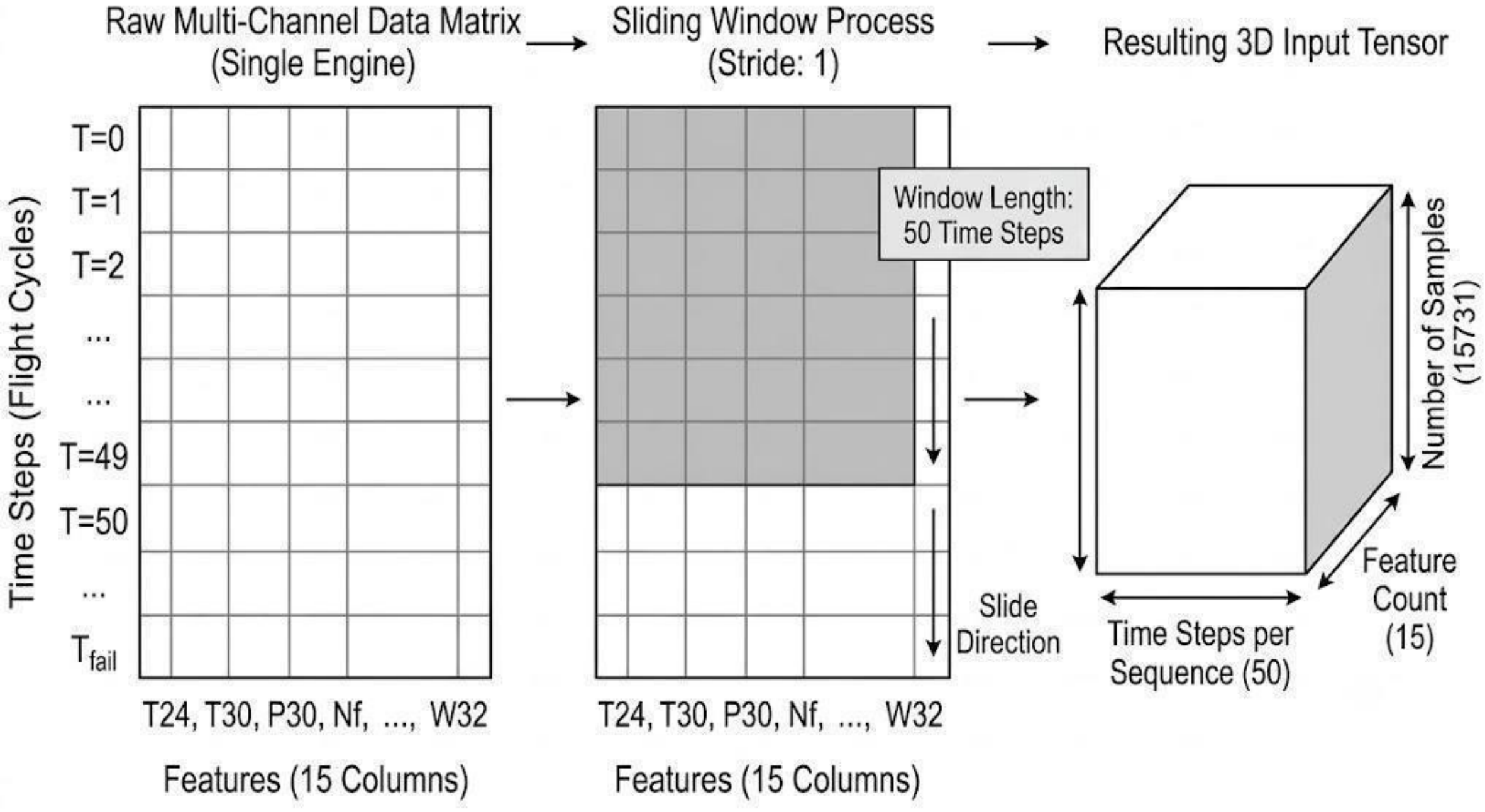

**Figure 3.** Schematic representation of the sliding window technique, illustrating the transformation of 2D multivariate sensor data into 3D tensor structures for LSTM sequential learning.

## 2.5. Deep Learning Architecture Design

A specialized two-layer LSTM architecture [23] was designed to execute the time series analysis and predict the continuous RUL values. The sequential flow of the network is structured as follows:

i. First Layer: An initial LSTM layer comprising 100 hidden units. This layer processes the incoming tensors, dimensioned at [50, 15] and [30, 24] for FD001 and FD004 respectively, and is configured to return the full sequence to feed the subsequent recurrent layer.
ii. Second Layer: A subsequent LSTM layer containing 50 hidden units. This layer is responsible for compressing the extracted temporal features into a dense representation.
iii. Output Layer: A fully connected dense layer with a single neuron and a Linear activation function, engineered to output the final continuous RUL prediction.

To systematically mitigate the risk of overfitting, which is a common challenge in deep learning models trained on finite datasets, dropout layers [24] were strategically integrated. A dropout rate of 20% was applied immediately after each LSTM layer, randomly deactivating neurons during each epoch to force the network to learn generalized patterns. The architecture was compiled using the Adam optimization algorithm [25], and the mean squared error (MSE) was selected as the loss function. Furthermore, an early stopping mechanism was implemented; if the model's validation loss failed to exhibit any improvement for 10 consecutive epochs, the training process was autonomously halted to preserve the optimal weights. The architectural design is intentionally streamlined; the entire model consists of 76.651 trainable parameters. This compact structure facilitates faster inference times and lower computational overhead compared to more parameter-heavy hybrid deep learning frameworks, ensuring its suitability for real-time what-if simulation deployment.

## 2.6. Performance Evaluation Metrics

To evaluate the predictive accuracy of the developed model on unseen test data, three distinct metrics were employed: the traditional RMSE, the $R^2$ score, and the domain-specific NASA asymmetric scoring function.

The RMSE is a standard metric for regression tasks, calculating the standard deviation of the prediction errors. The mathematical expression for RMSE is given in Eq. (3):

$$RMSE = \sqrt{\frac{1}{N}\sum_{i=1}^{N} (RUL_{Pred,i} - RUL_{Act,i})^2} \tag{3}$$

where N represents the total number of samples in the test set, the index i denotes each individual engine or flight cycle being evaluated, while $RUL_{Pred,i}$ and $RUL_{Act,i}$ correspond to the model's estimated remaining lifespan and the actual ground truth value for that specific instance, respectively.

Furthermore, the $R^2$ metric was incorporated to evaluate the model’s overall explanatory power. This metric represents the proportion of the variance in the target RUL that is predictable from the input sensor features, effectively demonstrating how well the model's predictions fit the real data. The mathematical expression for $R^2$ is given in Eq. (4):

$$R^2 = 1 - \frac{\sum_{i=1}^{N} (RUL_{Act,i} - RUL_{Pred,i})^2}{\sum_{i=1}^{N} (RUL_{Act,i} - \underline{RUL_{Act}})^2} \quad (4)$$

where $\underline{RUL_{Act}}$ represents the mean of the actual ground truth RUL values in the test set. The variables N, $RUL_{Act,i}$ and $RUL_{Pred,i}$ follow the same definitions as provided for Eq. (3).

While RMSE is an effective indicator of overall accuracy, it treats early and late predictions equally. However, in the aviation industry, overestimating an engine's remaining life (late maintenance) poses a catastrophic safety risk, potentially leading to in-flight failures. Conversely, underestimating the remaining life (early maintenance) only results in premature part replacement and financial loss. To account for this critical operational reality, the official NASA scoring function penalizes late predictions exponentially more heavily than early predictions.

Let $d_i$ represent the error margin, calculated as the difference between the prediction and the actual value as given in Eq. (5):

$$d_i = RUL_{Pred,i} - RUL_{Act,i} \quad (5)$$

The total asymmetric score S across all N test engines is computed using the mathematical formulation in Eq. (6):

$$S = \sum_{i=1}^{N} s_i \quad (6)$$

The penalty function $s_i$ for each individual prediction is defined as follows in Eq. (7) and Eq. (8):

$$s_i = e^{\frac{-d_i}{13}} - 1,\ if\ (d_i < 0) \quad (7)$$

$$s_i = e^{\frac{d_i}{10}} - 1,\ if\ (d_i \geq 0) \quad (8)$$

Through the application of this asymmetric penalty function, the model's performance was scrutinized from both a statistical perspective and an operational safety standpoint, prioritizing conservative estimations to mitigate the high-risk implications of delayed maintenance predictions.

### 2.7. Baseline Model for Benchmarking

To establish a robust performance baseline and properly justify the computational complexity of the proposed LSTM architecture, a traditional ensemble learning algorithm, RF [26], was trained on the exact same preprocessed dataset. Since traditional machine learning algorithms cannot natively process the 3-dimensional sequential tensors (samples × time steps × features) extracted via the sliding window technique, the data was mathematically flattened into a 2-dimensional tabular format prior to training. The RF model was configured with an ensemble of 100 decision trees to ensure algorithmic stability. This fundamental architectural difference allows for a fair and objective comparison of predictive capabilities between non-sequential standard machine learning logic and advanced sequential deep learning applied to aero-engine degradation [27].

Furthermore, to align with the current literature, the proposed architecture was also benchmarked against contemporary deep learning models, specifically BiLSTM and CNN-LSTM networks. The BiLSTM was included to evaluate whether processing degradation sequences in both directions provides prognostic advantages, while the CNN-LSTM was utilized to assess the impact of spatial feature extraction prior to temporal learning. All baseline deep learning models were trained using the exact same 3D tensor inputs and standardized hyperparameters to ensure a consistent and objective comparative analysis.

## 3. IMPLEMENTATION

The proposed framework was tested on independent test partitions of both the baseline FD001 and the multi-regime FD004 datasets, containing 100 and 248 completely unseen engine run-to-failure trajectories, respectively. The results are discussed not only in terms of statistical regression metrics but also regarding their practical implications for combat aircraft maintenance scheduling and flight safety.

### 3.1. Training Process and Model Convergence

The LSTM network was initially configured to train for a maximum of 100 epochs. As illustrated in the loss curves for the baseline FD001 dataset in Figure 4, the training and validation progression demonstrates the model's learning dynamics. The close proximity of the training (blue) and validation (red) curves throughout the process indicates a stable fit, effectively mitigating the risk of overfitting. Notably, following a rapid initial descent, a distinct secondary convergence drop is observed; the validation loss initiates this decline slightly before the 10th epoch, immediately followed by a corresponding drop in the training loss. This consistent and parallel decline indicates that the network generalizes effectively without memorizing the training data. To prevent the network from capturing structural noise during the extended epochs, an early stopping callback with a patience of 10 epochs dynamically monitored the validation loss. Consequently, the training phase was automatically halted at epoch 35, and the mechanism successfully restored the network's optimal weights from epoch 25. Furthermore, the integration of dropout layers, combined with the memory cells of the LSTM, enabled the network to process the raw, unsmoothed sensor signals while actively suppressing contextual noise.

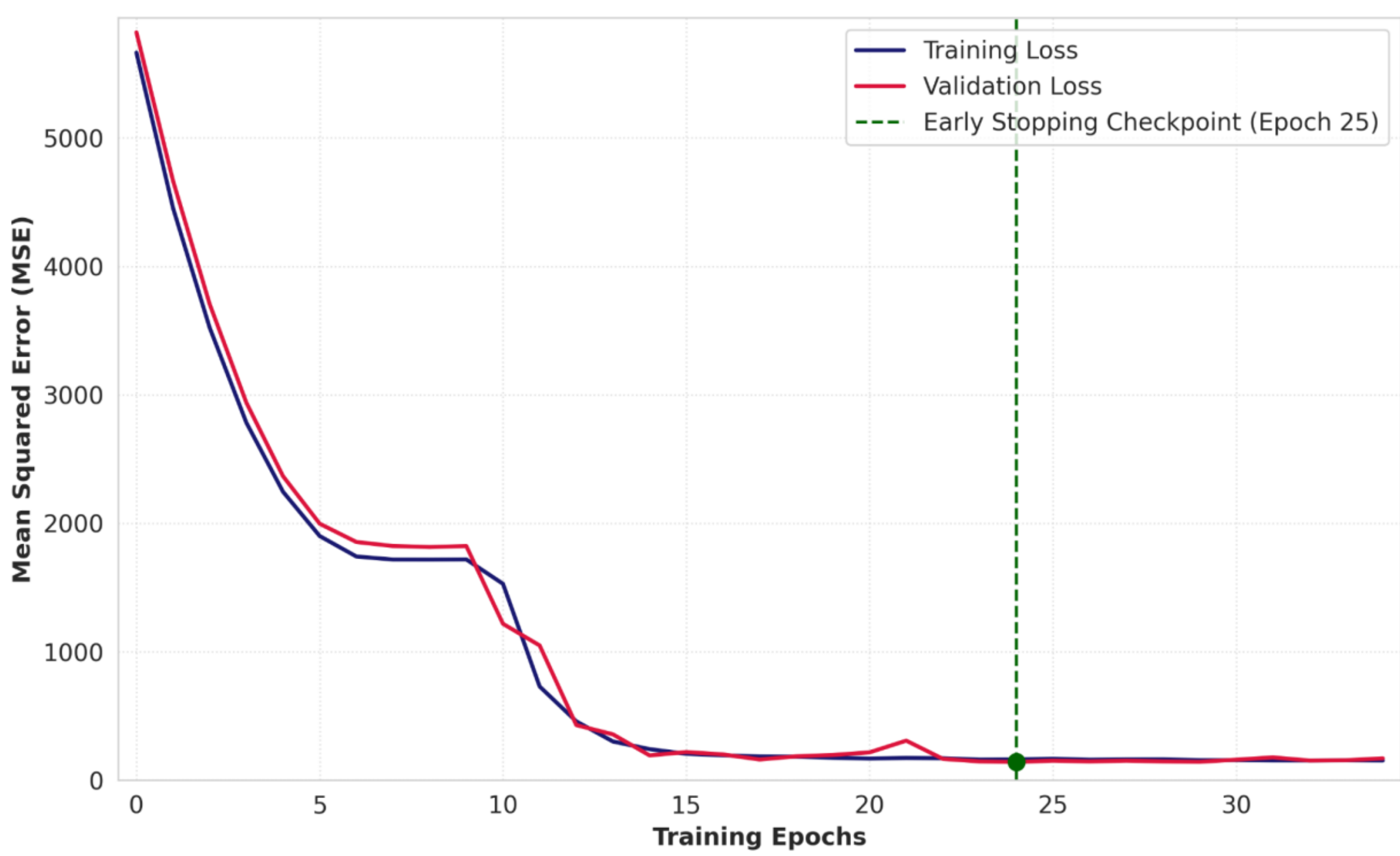

**Figure 4.** Convergence curves of training and validation loss for the proposed LSTM model. The loss is measured using mean squared error (MSE); the training process was automatically halted at epoch 35, with the optimal weights restored from epoch 25 by the early stopping mechanism.

## 3.2. Performance Metrics and Regression Analysis

To benchmark the efficacy of the autonomous feature extraction, three distinct evaluation metrics were computed on the test datasets. The results demonstrate that predictive accuracy can be achieved through deep learning, reducing the reliance on domain-specific manual feature engineering. By utilizing a multi-dimensional evaluation approach, the model's performance is assessed across both statistical precision and operational safety requirements. The metrics for the baseline FD001 dataset, highlighting the trade-off between error minimization and risk mitigation, are summarized in Table 2.

**Table 2.** Summary of the LSTM model's performance metrics on the NASA C-MAPSS FD001 test dataset.

| Evaluation Metric | Achieved Score | Operational Significance |
|---|---|---|
| RMSE | 13.28 | Measures standard deviation from the true RUL. |
| $R^2$ | 89.01% | Indicates the model's overall explanatory power. |
| NASA asymmetric score | 320.34 | Evaluates critical safety (penalizes late predictions). |

For the FD001 test set, the model achieved an RMSE of 13.28, establishing the baseline error margin for predicting the remaining useful life. Furthermore, the $R^2$ score reached 89.01%, indicating that the architecture accounts for a significant proportion of the variance within the complex degradation data. In terms of operational safety, the model recorded a NASA asymmetric score of 320.34, providing a quantifiable measure of its tendency to penalize late, high-risk predictions.

Following the baseline evaluation, the proposed regime-aware architecture was tested on the highly complex FD004 dataset, which comprises 248 test engines operating under six distinct flight regimes and two simultaneous fault modes. As expected in prognostic literature, predicting the RUL in a multi-regime environment inherently yields higher error margins compared to static conditions due to severe environmental noise. The performance metrics for the multi-regime dataset are summarized in Table 3.

**Table 3.** Summary of the LSTM model's performance metrics on the multi-regime NASA C-MAPSS FD004 test dataset.

| Evaluation Metric | Achieved Score | Operational Significance |
|---|---|---|
| RMSE | 15.71 | Measures standard deviation in a multi-regime environment. |
| $R^2$ | 86.64% | Indicates the model's explanatory power under varying flight conditions. |
| NASA asymmetric score | 1533.89 | Evaluates critical safety within complex, dual-fault scenarios. |

Despite the extreme volatility of the FD004 dataset, the model achieved an RMSE of 15.71 and maintained an $R^2$ score of 86.64%. The NASA asymmetric score of 1533.89 is structurally higher than that of the baseline dataset, reflecting the cumulative penalty across a significantly larger test partition (248 engines vs. 100 engines) and the inherent difficulty of dual-fault prognostics. However, these results validate that the integration of K-Means clustering with localized Z-score standardization effectively enables the LSTM network to track mechanical wear even through severe operational shifts.

Beyond standard accuracy metrics, the distribution of prediction errors (residuals) was analyzed to evaluate the model's statistical unbiasedness. As illustrated in Figure 5, which displays the error margin histogram, the residuals form a near-symmetrical normal distribution centered tightly around the zero-error line. Specifically, the model exhibits a virtually negligible residual mean error (Bias) of only -0.12 flight cycles, alongside a standard deviation (SD) of $\pm$ 13.28 cycles. The fact that the standard deviation is mathematically equivalent to the overall RMSE (13.28) is a highly significant indicator; it validates that the prediction error is almost entirely driven by natural dataset variance rather than systematic bias ($RMSE^2 = Bias^2 + SD^2$). This near-zero bias and symmetrical distribution mathematically prove that the LSTM model is reliable; it does not suffer from systematic early or late prediction tendencies, representing a critical safety factor for aerospace applications.

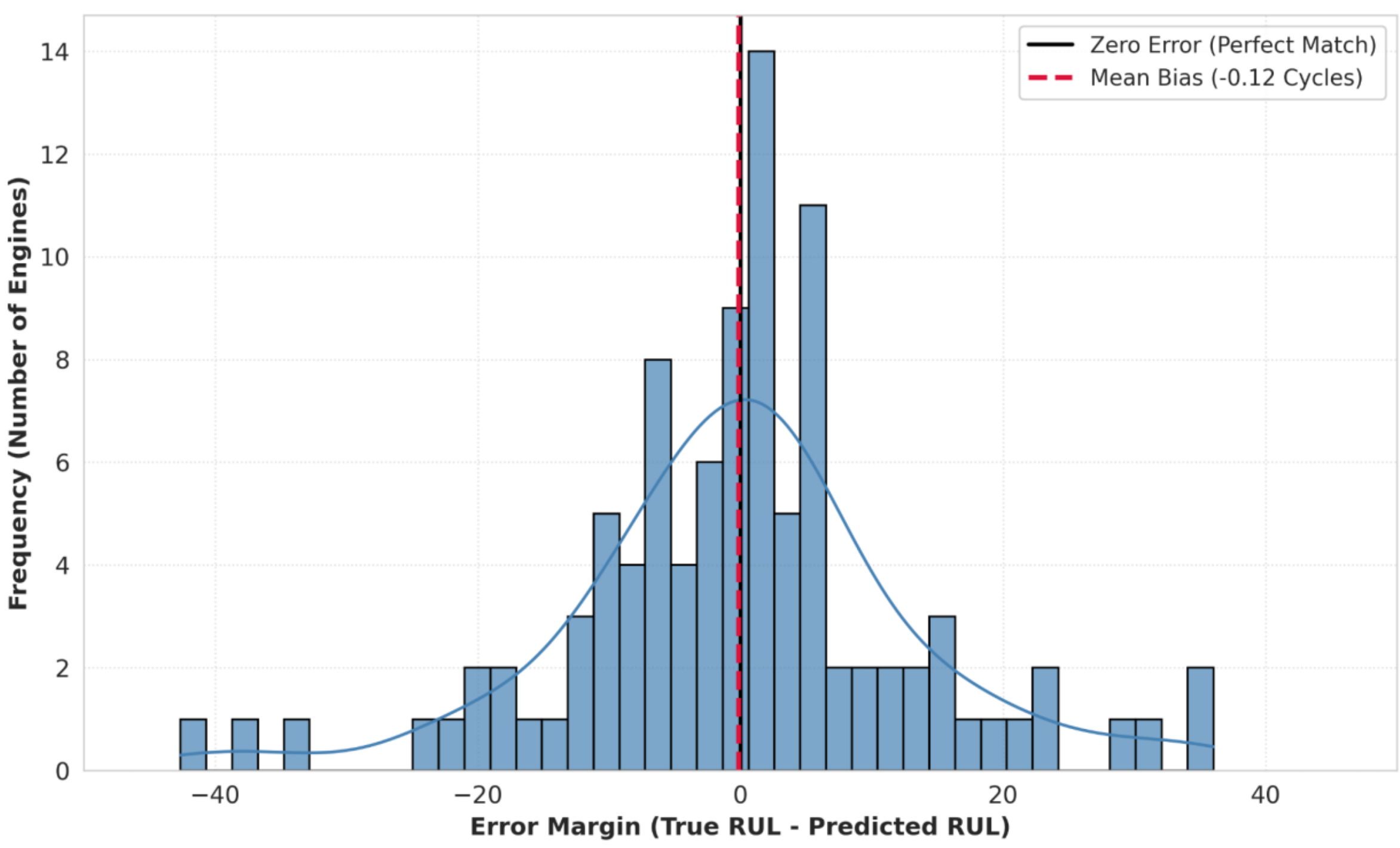

**Figure 5.** Prediction error (residual) distribution of the baseline LSTM model, demonstrating a near-perfect symmetrical distribution centered around a minimal bias of -0.12 cycles.

To visualize the model's predictive tracking capability across varying stages of engine degradation, Figure 6 presents a comparative plot of the true RUL versus the LSTM predictions for all test engines in the FD001 dataset. The engines are sorted in ascending order based on their true RUL values to illustrate the prediction trend clearly. The plot demonstrates that the LSTM predictions align closely with the actual RUL when the remaining life is low (left side of the graph), which corresponds to the primary degradation phase for maintenance intervention. Conversely, for healthier engines with higher RUL values (right side of the graph), the prediction variance naturally increases. Notably, in these higher ranges, the model generally tends to underestimate rather than overestimate the RUL. This statistical behavior corresponds with the asymmetric penalty structure of the NASA scoring function, exhibiting a risk-averse predictive tendency that favors preemptive maintenance scheduling over delayed intervention.

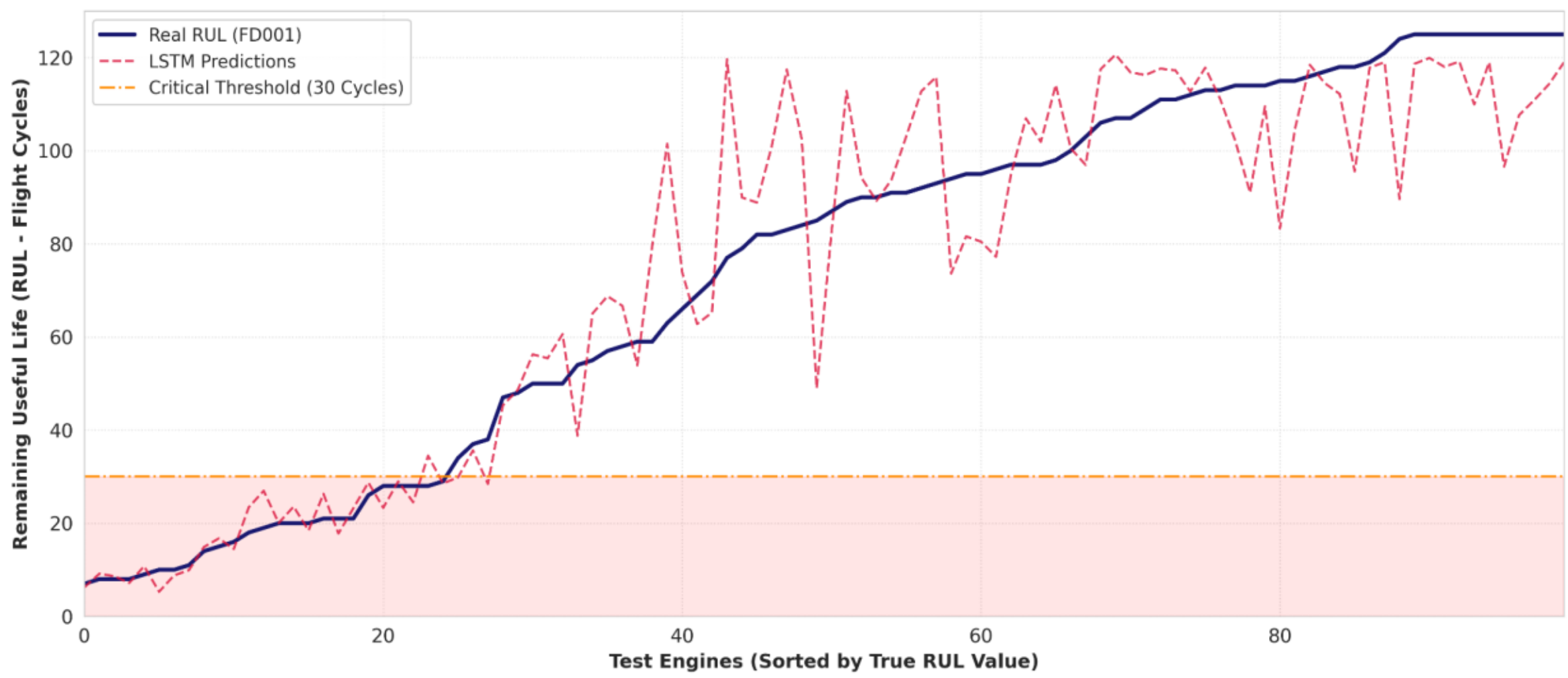

**Figure 6.** Comparison of True RUL values against LSTM predictions on the NASA C-MAPSS FD001 test dataset.

To further validate the model's prognostic robustness under severe operational noise, Figure 7 illustrates the comparative RUL tracking for all 248 test engines in the multi-regime FD004 dataset. Despite the extreme volatility introduced by six flight regimes and dual fault modes, the LSTM architecture successfully captures the overall degradation trajectory. Notably, as the engines approach the critical 30-cycle early-warning threshold (indicated by the shaded danger zone), the predictions tightly converge with the ground truth. This precise alignment in the terminal phase confirms that the regime-aware normalization and autonomous feature extraction pipeline effectively filters out environmental noise precisely when maintenance intervention is most crucial.

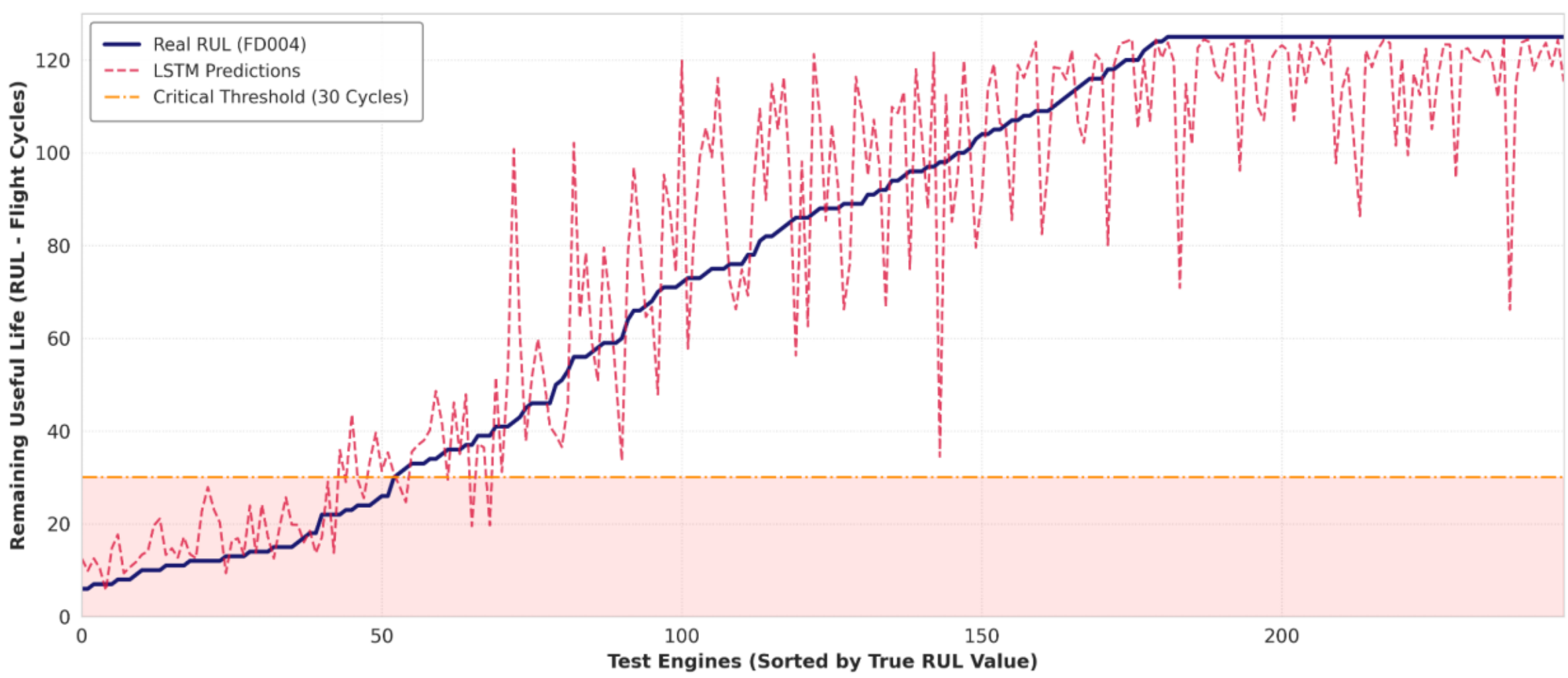

**Figure 7.** Comparison of True RUL values against LSTM predictions for the multi-regime NASA C-MAPSS FD004 test dataset, highlighting the predictive alignment within the critical 30-cycle safety threshold.

A comparative analysis was conducted to evaluate the LSTM architecture in handling sequential degradation data. The proposed model was first benchmarked against a traditional RF baseline. The results, presented in Table 4, show that the deep learning approach achieved better performance across the evaluated metrics, particularly in reducing the safety-critical NASA asymmetric score.

**Table 4.** Performance comparison between the proposed LSTM architecture and the RF baseline.

| Model | RMSE | NASA Score | $R^2$ |
|---|---|---|---|
| RF | 15.54 | 399.01 | 84.96% |
| Proposed LSTM | 13.28 | 320.34 | 89.01% |

As detailed in Table 4, the proposed LSTM model outperforms the traditional RF baseline across all critical metrics. Specifically, the LSTM network achieved a lower RMSE of 13.28 compared to the RF's 15.54 (a reduction of 2.26 cycles), while improving the $R^2$ value by 4.05%. Most notably, the deep learning architecture effectively reduced the safety-critical NASA asymmetric score, dropping from 399.01 to 320.01. This substantial reduction in the penalty score highlights the model's enhanced capability in minimizing high-risk, late predictions. It demonstrates the advanced capacity of the LSTM to mitigate severe prediction risks, thereby justifying its implementation for aviation prognostics. To further position the study relative to current methodologies, the proposed architecture was additionally benchmarked against two contemporary deep learning approaches: BiLSTM and CNN-LSTM. All models were evaluated under standardized training parameters to ensure consistency.

**Table 5.** Performance comparison of the proposed architecture against contemporary deep learning baselines on the baseline FD001 dataset.

| Model | RMSE |
|---|---|
| CNN-LSTM | 15.62 |
| BiLSTM | 14.44 |
| Proposed LSTM | 13.28 |

As shown in Table 5, the BiLSTM and CNN-LSTM models yielded RMSE scores of 14.44 and 15.62, respectively. While these architectures are widely utilized in prognostic tasks, they did not outperform the proposed streamlined LSTM despite their greater architectural complexity. Specifically, the CNN-LSTM hybrid, despite its spatial feature extraction capabilities, resulted in a higher error margin compared to the standard unidirectional LSTM for this specific multivariate sensor structure. Overall, the comparative evaluation demonstrates that the proposed LSTM architecture provides an optimal balance, achieving the lowest RMSE of 13.28 while maintaining architectural simplicity and computational efficiency. This indicates that for the FD001 dataset, the proposed model offers superior predictive precision without the overhead associated with more complex hybrid frameworks.

While the primary optimized model achieved a superior RMSE of 13.28, it is well-documented that deep learning architectures are inherently sensitive to random weight initializations during non-convex optimization. To ensure statistical stability and confirm that the reported predictive capabilities are not merely an artifact of favorable weight initialization, a robustness analysis was conducted across multiple independent training iterations using varying random seeds. As detailed in Table 6, across five successfully converged runs, the evaluations yielded a mean RMSE of 13.86 with a standard deviation of $\pm$0.58. This steady-state average demonstrates that even under unconstrained initial conditions, the proposed LSTM framework consistently outperforms not only the traditional RF baseline but also the more complex BiLSTM and CNN-LSTM architectures. This low variance confirms the stable learning capability and operational reliability of the network for aviation prognostics. It should be noted that the 13.06 RMSE represents the best seed result, whereas 13.28 remains the primary optimized model result used for the main comparison.

**Table 6.** Robustness analysis and variability of the proposed LSTM architecture across multiple independent training runs.

| Training Iteration | RMSE Score | Mean (Average) | Standard Deviation |
|---|---|---|---|
| Run 1 | 14.82 | 13.86 | $\pm$ 0.58 |
| Run 2 | 13.06 | | |
| Run 3 | 14.07 | | |
| Run 4 | 13.54 | | |
| Run 5 | 13.80 | | |

## 3.3. Asymmetric Safety Scoring and Discussion

While RMSE is a standard evaluation metric, its symmetric nature treats early and late predictions equally. In the context of military aviation, overestimating an engine's remaining lifespan (late prediction) introduces severe flight safety risks, whereas underestimating it (early prediction) primarily results in premature maintenance costs. The NASA asymmetric score function mathematically addresses this operational reality by heavily penalizing late predictions. As demonstrated in the results, the developed LSTM model achieved a cumulative NASA Score of 320.34 across the baseline FD001 test engines, and 1533.89 on the highly volatile multi-regime FD004 dataset. While the FD004 score is structurally higher due to the larger test population and severe operational noise, both evaluations indicate that the algorithm exhibits a risk-averse predictive tendency. The network effectively minimizes dangerous overestimations across varying flight conditions, supporting proactive maintenance scheduling to mitigate the risk of critical in-flight failures in combat aircraft.

### 3.4. Critical Failure Safety Analysis

Although the primary objective of the deep learning network is to output continuous numerical RUL values (regression), real-world operational protocols require distinct, actionable intervention triggers. Therefore, an RUL threshold dropping below 30 flight cycles was categorized as critical maintenance (danger). This 30-cycle threshold represents a realistic logistical lead time required to schedule maintenance, procure spare parts, and safely ground the aircraft without disrupting mission readiness. To evaluate the model's performance in this binary context (healthy vs. critical), a confusion matrix and a receiver operating characteristic (ROC) curve were generated, as shown in Figure 8 and Figure 9, respectively. To ensure that the classification metrics are robust and not merely a byproduct of class imbalance, the exact class distribution of the test set was analyzed. Out of the 100 total test engines, the distribution for the 30-cycle threshold comprises 75 healthy engines (>30 cycles) and 25 critical engines (≤30 cycles), providing sufficient representation for the critical minority class. Furthermore, to explicitly justify the 30-cycle selection, a sensitivity analysis was conducted by evaluating the model across 20, 30, and 40-cycle boundaries, as detailed in Table 7. As shown in Table 7, the evaluation of the ROC curve for the baseline 30-cycle threshold yielded an area under the curve (AUC) value of 0.9973. The confusion matrix further corroborates this performance, displaying a high number of correct classifications across the test dataset (73 true negatives and 24 true positives). This translates to an overall classification accuracy of 97.00%, with a precision of 92.31%, a sensitivity of 96.00% and a specificity of 97.33%. In practical terms, based on the test set distribution, the system demonstrates a suitably conservative false-positive rate. This minimizes the likelihood of premature groundings, thereby reducing unnecessary maintenance costs and effectively supporting overall fleet availability. The limited number of misclassifications (specifically, 1 false negative and 2 false positives) corresponds to instances where the engine was at the borderline of the critical zone, but the continuous regression model estimated it slightly above or below the strict threshold (e.g., predicting 31 cycles while the ground truth is 29). Given the continuous nature of the RUL regression, these borderline cases are mathematically expected and can be effectively managed by defining a narrow safety buffer (e.g., ±5 cycles) in real-world deployment. Overall, the ROC-AUC and confusion matrix metrics confirm that the proposed LSTM architecture is not only a precise regression tool but also a reliable binary safety classifier for critical mission planning.

**Table 7.** Classification performance under different RUL-based early-warning thresholds.

| Threshold (Flight Cycles) | Healthy | Critical | AUC | Precision | Sensitivity | Specificity |
|---|---|---|---|---|---|---|
| 20 Cycles | 84 | 16 | 0.9911 | 0.9333 | 0.8750 | 0.9881 |
| 30 Cycles (Baseline) | 75 | 25 | 0.9973 | 0.9231 | 0.9600 | 0.9733 |
| 40 Cycles | 72 | 28 | 1.0000 | 0.9655 | 1.0000 | 0.9861 |

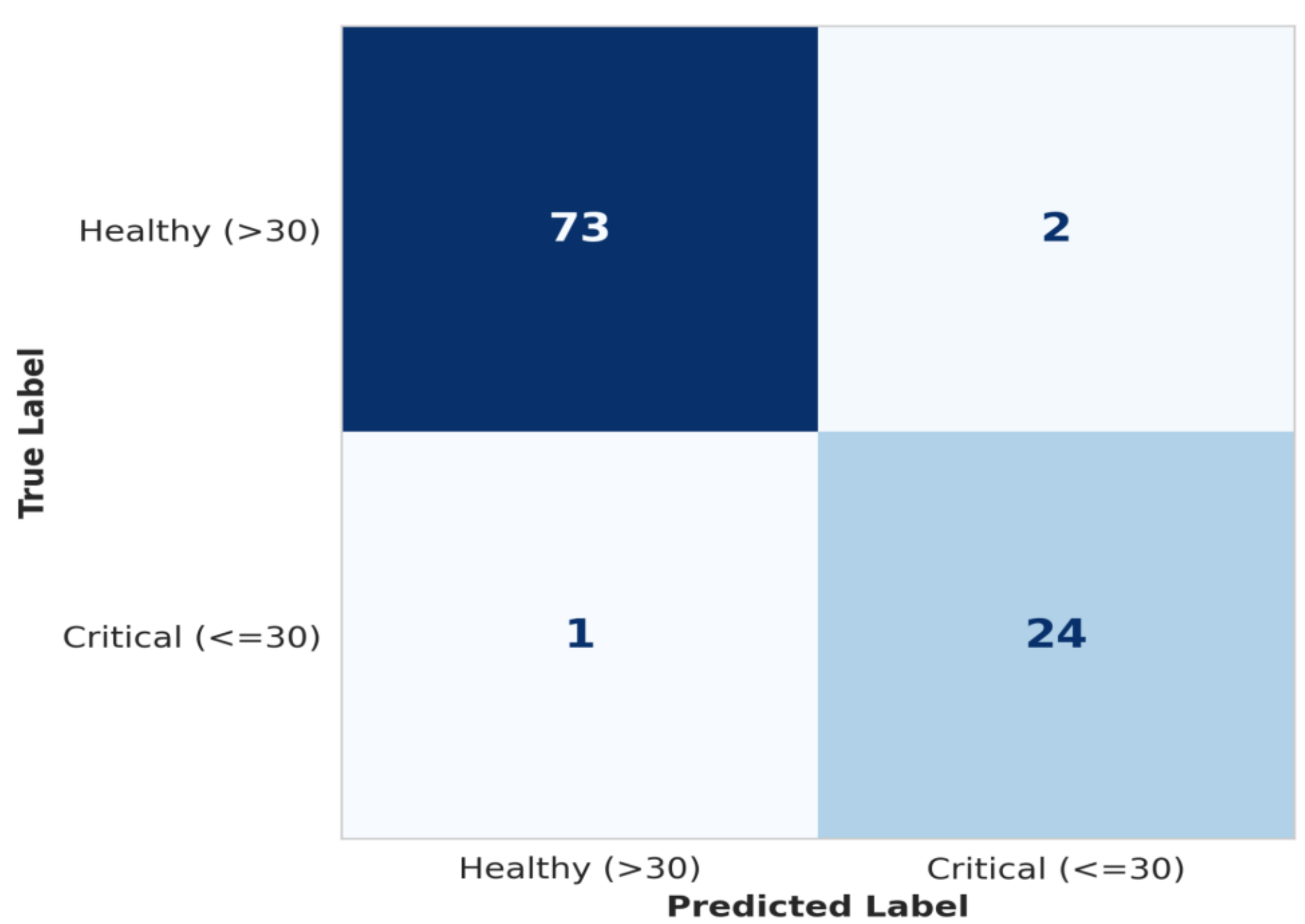

**Figure 8.** Maintenance decision confusion matrix (Threshold: ≤ 30 cycles).

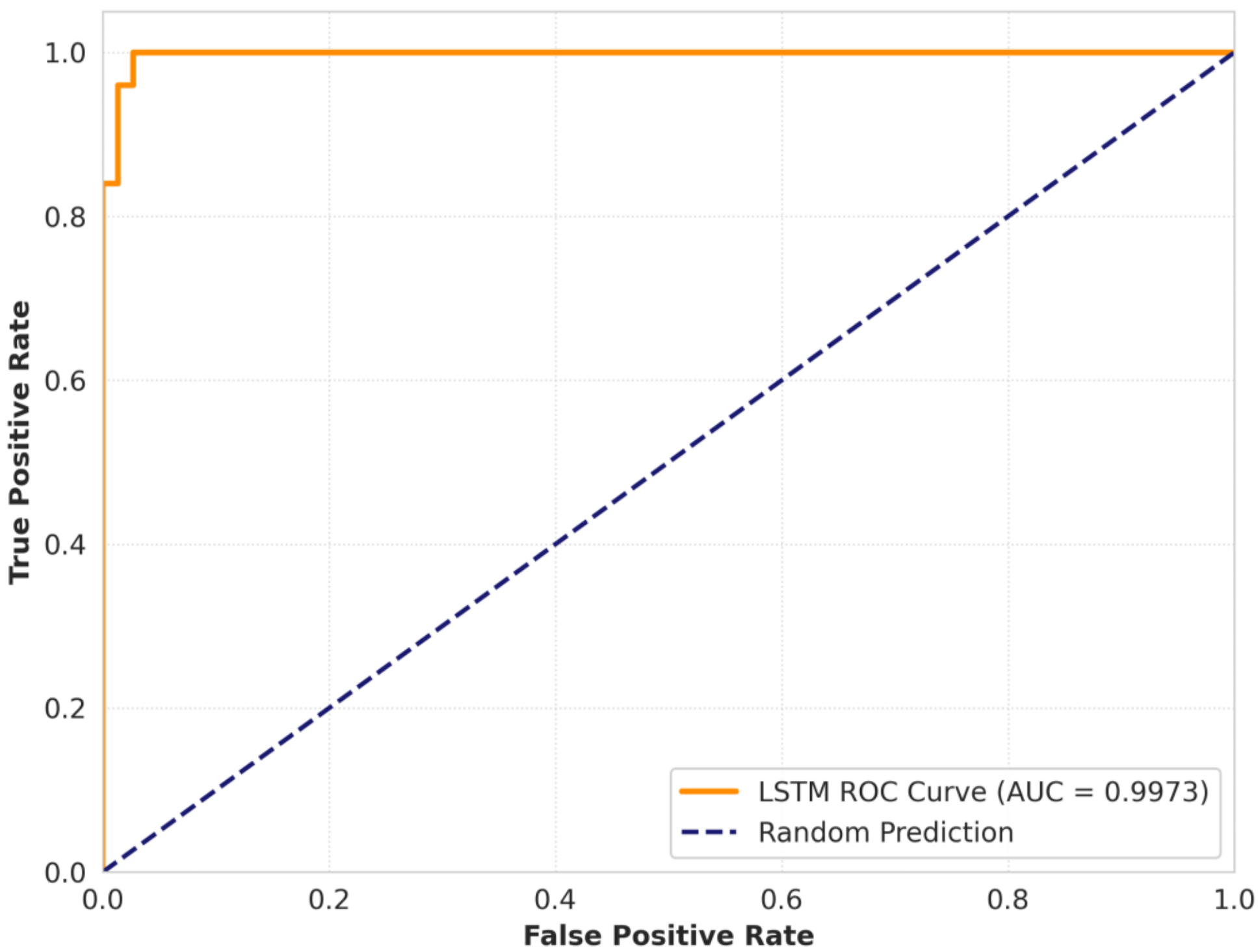

**Figure 9.** ROC curve and AUC for critical engine failure detection.

### 3.5. Interactive Decision-Support What-If Simulator

To bridge the gap between theoretical deep learning metrics and operational fleet management, the proposed architecture was operationalized into an interactive decision-support what-if simulator. As noted in recent prognostic literature, while data-driven models effectively extract degradation patterns from multivariate time-series arrays, their raw numerical outputs require contextualization for mission planning. This simulator functions as a forward-looking exploratory framework, allowing flight commanders to dynamically assess the impact of variable operational stress on the remaining useful life of combat aircraft engines. To ensure the scientific validity of this what-if analysis, the simulation framework is intentionally built upon the FD001 test partition. Because FD001 represents a controlled environment with a single operating condition, it provides an uncorrupted baseline. If an artificial stress factor is applied to a multi-regime dataset such as FD004, it becomes significantly difficult to distinguish whether the resulting RUL degradation stems from simulated combat stress or from structural environmental changes (such as sudden changes in altitude or Mach number). Thus, utilizing the FD001 dataset ensures a methodologically sound experimental design free from confounding variables.

The core mechanism of the simulator relies on real-time tensor manipulation. Through a graphical user interface (GUI), the system displays a static global base RMSE to maintain the scientific context of the model's overall error margin, alongside the baseline RUL of the selected engine under standard operating conditions. Operators can select from the entire evaluation fleet of 100 independent test engines and apply a continuous "Operational Stress Multiplier" (ranging from 0.8x to 1.5x) to the baseline sensor sequences, which triggers the system to output a dynamically updated current RUL. To prevent this from being a naive scalar transformation, the system incorporates a structured thermodynamic sensitivity logic. The 21 input sensors are categorized into four distinct sensitivity tiers based on their physical placement within the turbofan engine. For instance, primary core components (e.g., high-pressure compressor, combustor, and high-pressure turbine) form the highest sensitivity group, exhibiting immediate anomalous behaviors at minor stress increments (≥1.05x). Conversely, peripheral components like the fan and bypass operate with higher thermal inertia, showing degradation signatures only under extreme stress profiles (≥1.45x). When the user adjusts the stress parameter, the system dynamically applies these tiered thermodynamic deviations to the baseline (1, 50, 15) input tensors. To accurately model cascading degradation without distorting the inverse correlations of the normalized sensor data, the LSTM's baseline sequential inference is mathematically coupled with an exponential thermodynamic wear factor, instantly calculating the structurally adjusted dynamic RUL.

The operational utility and dynamic responsiveness of this decision-support framework are demonstrated through a spectrum of three distinct what-if scenarios. First, to evaluate a conservative flight profile, Figure 10 illustrates test engine #4 under a reduced stress multiplier of 0.95x. This scenario mimics optimal cruising conditions with lowered thermodynamic loads. Under baseline conditions, the engine possessed an RUL of 90.9 cycles. The application of the 0.95x multiplier mathematically alleviates the simulated sensor strain, dynamically extending the forecasted RUL to 103.3 cycles. The visual diagnostic panel confirms this stable state by displaying a "Healthy" diagnostic status, visually retaining all sensor tiers in a safe green state to indicate the absence of any anomalous deviations.

Second, to evaluate a mild operational load, Figure 11 presents test engine #11 under a 1.05x combat stress profile. Under standard conditions, the LSTM model predicted an RUL of 96.9 flight cycles. The 1.05x multiplier introduces elevated core temperatures and pressures, subsequently reducing the dynamic RUL to 85.7 cycles. The visualization panel effectively captures this localized stress by updating the diagnostic status to an "Elevated Core Stress" warning, visually flagging the core thermodynamic sensors (e.g., T30, Ps30, W31) as active orange stress points while the peripheral sensors remain within safe operational margins.

Finally, to evaluate severe mission limitations, Figure 12 presents test engine #67 under an extreme combat stress multiplier of 1.45x. The engine possessed a robust baseline RUL of 121.6 cycles. Under the simulated aggressive flight maneuvers and maximum thrust demands, the system modeled widespread cascading degradation across all four sensitivity groups. The LSTM model captured this severe multivariate distortion, recalculating the RUL dynamically down to 48.0 cycles. Because the forecasted value remains above the strict 30-cycle early-warning threshold, the engine avoids a critical grounding status. However, the diagnostic logic dynamically updates to a "Severe

Thermodynamic Stress" warning, visually converting the entire sensor array to a critical red status to explicitly map the structural cost of the aggressive mission.

Ultimately, this decision-support simulator enables maintenance crews to mathematically quantify the exact structural impact of various military operations. By transforming static deep learning regressions into dynamic operational intelligence, the framework supports proactive fleet readiness and optimal maintenance scheduling without risking physical hardware in extreme flight profiles.

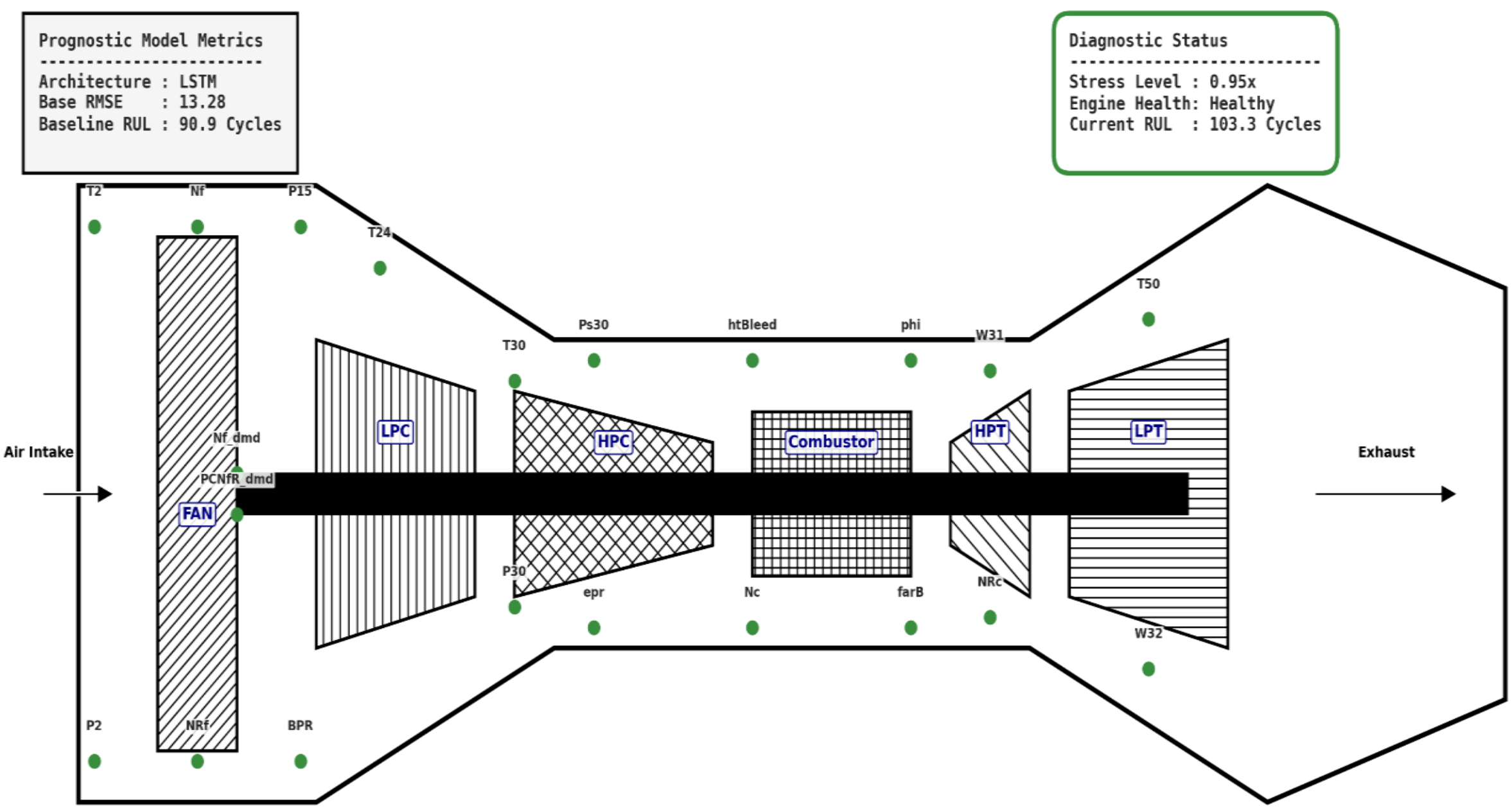

**Figure 10.** What-if analysis results for Test Engine #4. Under a simulated conservative 0.95x operational stress multiplier, the predicted RUL dynamically extends from a baseline of 90.9 cycles up to 103.3 cycles. The diagnostic panel visually retains a safe green "Healthy" status, indicating reduced thermodynamic load.

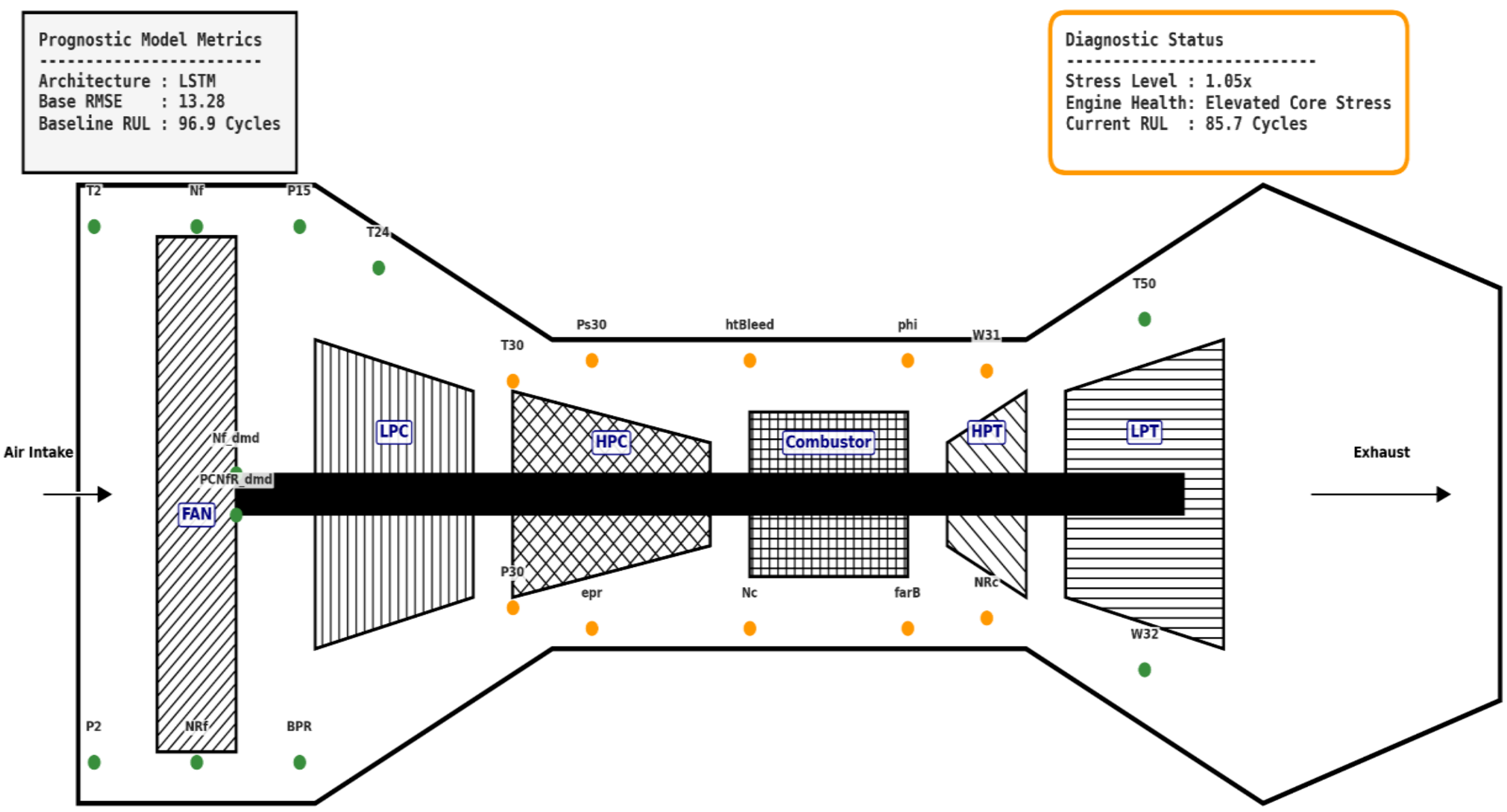

**Figure 11.** What-if analysis results for Test Engine #11. Under a simulated 1.05x operational stress multiplier, the predicted RUL dynamically adjusts from 96.9 to 85.7 cycles. Core sensors indicate localized thermodynamic stress, dynamically visualized with orange warning flags.

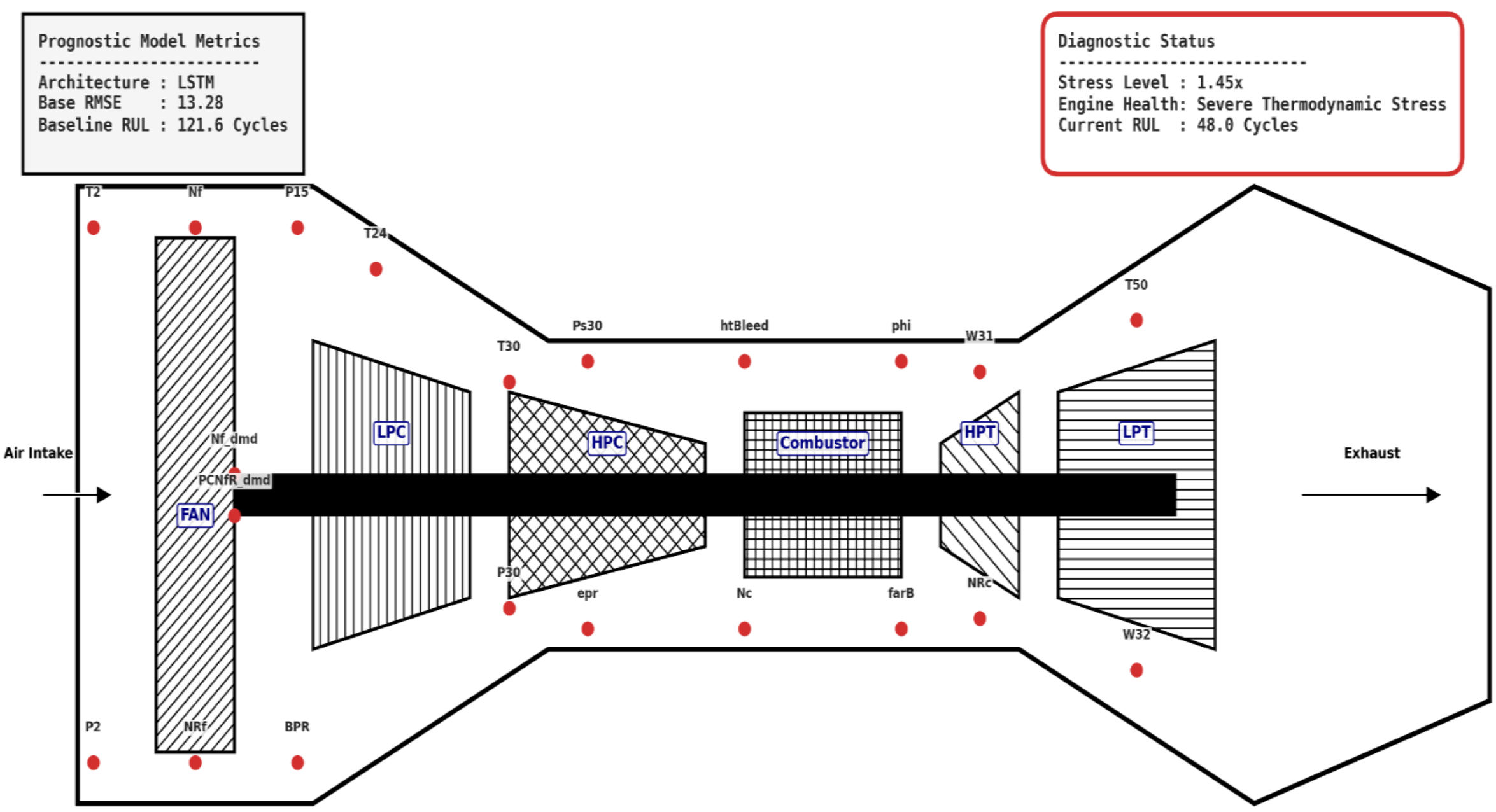

**Figure 12.** What-if analysis results for Test Engine #67 under an extreme combat stress profile (1.45x multiplier). The simulator models widespread cascading sensor deviation, resulting in a forecasted RUL reduction from 121.6 down to 48.0 flight cycles, explicitly mapped with a critical red diagnostic status.

### 3.6. Future Work

While the proposed decision-support simulation framework establishes a solid, data-driven foundation for aero-engine prognostics, several critical paths emerge for future research to extend its operational impact within military aviation. First, to bridge the gap between data-driven heuristics and enhanced thermodynamic fidelity, future iterations will explore the integration of physics-informed neural networks (PINNs). Incorporating explicit aerothermal constraints and physical laws directly into the deep learning architecture will further enhance the model's reliability under extreme combat profiles. Second, researching computationally optimized self-attention mechanisms and Transformer encoders could improve the extraction of long-range temporal dependencies across diverse sensor channels, provided that real-time inference efficiency is strictly maintained. Finally, future studies aim to transition this framework from a simulated environment into a hardware-in-the-loop (HIL) testing phase, ultimately paving the way for its deployment as a real-time, on-board predictive maintenance advisor that actively supports fleet readiness and mission safety.

## 4. CONCLUSION

In this paper, an autonomous predictive maintenance framework aligned with Industry 4.0 paradigms was developed to supplement traditional time-based aircraft maintenance strategies. By leveraging a streamlined, deep learning-based LSTM architecture, the complex problem of RUL estimation for turbofan engines was effectively addressed. Empirical evaluations on the NASA C-MAPSS datasets validated the methodology's superiority. On the baseline FD001 dataset, the model achieved an RMSE of 13.28, an $R^2$ score of 0.8901, and a safety-critical NASA asymmetric penalty score of 320.34. These metrics demonstrate enhanced predictive accuracy and penalty minimization compared to traditional machine learning baselines and contemporary deep learning frameworks, including BiLSTM and CNN-LSTM architectures. Furthermore, the model exhibited high architectural robustness, yielding a stable mean RMSE of 13.86 ($\pm$0.58) across multiple independent training iterations.

A primary contribution of this study is the model's generalizability across complex operational environments. Utilizing a regime-aware K-Means clustering and intra-regime Z-score standardization pipeline, the framework successfully adapted to the highly volatile, multi-fault FD004 dataset, achieving an RMSE of 15.71. This confirms that the proposed preprocessing and modeling pipeline handles severe flight condition variations and structural noise effectively, without requiring architectural modifications.

Moving beyond standard regression tasks, this research introduced two operational implementations to bridge the gap between theoretical prognostics and real-world applications. First, continuous predictions were transformed into a binary failure detection classifier, achieving an AUC of 0.9973 at a critical 30-cycle threshold with a minimal false-positive rate. Second, an interactive decision-support what-if simulator was developed, allowing flight commanders to dynamically quantify the structural cost of variable combat stress multipliers on engine lifespan. To support open science, the complete Python source code, preprocessing pipeline, and simulator will be made publicly available upon publication. Ultimately, this framework provides a highly reliable, data-driven solution for defense and commercial aviation agencies to minimize unscheduled downtime, optimize maintenance scheduling, and ensure operational mission safety.

## VITAE

**Fatih Ürgen** received his B.Sc. degree in Physics Engineering from Faculty of Engineering and Natural Sciences, Istanbul Medeniyet University (IMU), Türkiye in 2025. He is currently pursuing a double major (B.Sc.) in Electrical and Electronics Engineering from IMU, and an M.Sc. degree in Physics Engineering at Istanbul Technical University (ITU), Türkiye.

**Doğay Altınel** received his B.Sc. degree in Electrical and Electronics Engineering from Faculty of Engineering, Hacettepe University, Ankara, Türkiye in 1992. He received his M.Sc. and Ph.D. degrees in Telecommunications Engineering from Istanbul Technical University (ITU), Türkiye, in 2014 and 2019, respectively. He is currently working as an Assistant Professor at the Department of Electrical and Electronics Engineering, Faculty of Engineering and Natural Sciences, Istanbul Medeniyet University (IMU), Türkiye.